%% file: main-icaps-26.tex
\documentclass[letterpaper]{article} 
\usepackage[]{aaai2026}  
\usepackage{times}  
\usepackage{helvet}  
\usepackage{courier}  
\usepackage[hyphens]{url}  
\usepackage{graphicx} 
\usepackage{natbib}  
\usepackage{caption} 
\usepackage{algorithm}
\usepackage[noend]{algorithmic}

\usepackage{amsmath}
\usepackage{amsthm}
\usepackage{booktabs}
\usepackage{enumitem}
\usepackage{graphicx}
\usepackage{color}
\usepackage{cleveref}
\usepackage{float}
\usepackage{mathtools}
\usepackage{makecell}
\usepackage{multirow}

\usepackage{gr-macros}

\usepackage[textsize=tiny
            ]{todonotes}

\usepackage{subcaption} 

\usepackage{newfloat}
\usepackage{listings}
\DeclareCaptionStyle{ruled}{labelfont=normalfont,labelsep=colon,strut=off} 
\floatstyle{ruled}
\newfloat{listing}{tb}{lst}{}
\floatname{listing}{Listing}
\title{Removing Planner Bias in Goal Recognition Through Multi-Plan Dataset Generation}

\author{
    Mustafa F. Abdelwahed\textsuperscript{1},
    Felipe Meneguzzi\textsuperscript{2,3}
    Kin Max Piamolini Gusmão\textsuperscript{2},
    Joan Espasa\textsuperscript{1}
}
\affiliations{
    \textsuperscript{\rm 1} University of St Andrews, School of Computer Science, UK\\
    \texttt{\{ma342, jea20\}@st-andrews.ac.uk}\\
    \textsuperscript{\rm 2} Pontifical Catholic University of Rio Grande do Sul, Brazil\\
    \texttt{Kin.Gusmao@edu.pucrs.br}\\
    \textsuperscript{\rm 3} University of Aberdeen, UK\\
    \texttt{felipe.meneguzzi@abdn.ac.uk}
}

\usepackage{bibentry}

\newtheorem{definition}{Definition}

\begin{document}

\maketitle

\begin{abstract}
Autonomous agents require some form of goal and plan recognition to interact in multiagent settings. Unfortunately, all existing goal recognition datasets suffer from a systematical bias induced by the planning systems that generated them, namely heuristic-based forward search. 
This means that existing datasets lack enough challenge for more realistic scenarios (e.g., agents using different planners), which impacts the evaluation of goal recognisers with respect to using different planners for the same goal. 
In this paper, we propose a new method that uses top-k planning to generate multiple, different, plans for the same goal hypothesis, yielding benchmarks that mitigate the bias found in the current dataset. 
This allows us to introduce a new metric called Version Coverage Score (VCS) to measure the resilience of the goal recogniser when inferring a goal based on different sets of plans. 
Our results show that the resilience of the current state-of-the-art goal recogniser degrades substantially under low observability settings.
\end{abstract}


\input{sections/01-introduction}

\input{sections/02-motivating-example}
\input{sections/03-background}
\input{sections/04-resilience-idea}
\input{sections/05-exps}
\input{sections/06-conclusion-future-work}

\newpage
\bibliography{smt-gr}


\end{document}

%% file: sections/01-introduction.tex
\section{Introduction}

Goal and plan recognition are the related tasks of inferring, from observations of an agent's behaviour, the agent's goal and the plan it is following towards such goal.  
The ability to carry out such recognition is critical when dealing with agents operating in open multiagent environments for two main reasons. 
First, it allows one to anticipate an agent's behaviours, and thus reasoning about positive or negative interactions between different agents sharing the same environment. 
Second, it allows agents to coordinate their behaviour without explicit communication (e.g., to avoid negative interference between potentially conflicting plans). 

Following the seminal approach by \citet{RamirezGeffner2009}, recent research yielded substantial progress on methods for goal and plan recognition in single agent settings~\cite{RamirezGeffner2009,RamirezG_AAAI2010,SukthankarGoldmanGeibEtAl2014,MartinMorenoSmith2015,SohrabiRiabovUdrea2016,RamonNirMeneguzzi_AAAI2017,VeredKaminka2017,PereiraOrenMeneguzzi2020,RdeASantosMeneguzziPereiraEtAl2021,AmadoMirskyMeneguzzi2022,SerinaChiariGereviniEtAl2025}.
These methods vary considerably, from relying on a full planning algorithm to find observation compliant plans~\cite{RamirezG_AAAI2010,SohrabiRiabovUdrea2016,VeredKaminka2017}, to computing heuristic estimates of such plans~\cite{MartinMorenoSmith2015,RamonNirMeneguzzi_AAAI2017,PereiraOrenMeneguzzi2020}, to building machine-learning models of the agent preferences~\cite{AmadoMirskyMeneguzzi2022,ChiariGereviniPercassiEtAl2023,SerinaChiariGereviniEtAl2025}.
Indeed, most recent approaches achieve substantial accuracy, often at low total computational cost~\cite{PereiraOrenMeneguzzi2020,RdeASantosMeneguzziPereiraEtAl2021}. 
Common in virtually all such approaches is their use of either the problem set from \citet{RamirezG_AAAI2010}, or a more recent problem set \cite{ramon_fraga_pereira_2017_825878} derived from~\citet{PereiraOrenMeneguzzi2020}.
Indeed, they were all created by planning systems that rely on heuristic search~\cite{GereviniSaettiSerina2003,Helmert2006,RichterWestphal2010}, and where systematic biases (for instance, sub-optimality) is a result of an inadmissible heuristic~\cite{RichterWestphal2010}, an inadmissible search procedure (e.g., GBFS or weighted A$^*$ in \cite{Helmert2006}), or sampling from top-k plans~\cite{KatzSohrabiUdreaEtAl2018,GereviniSaettiSerina2003}. 

Current goal recognition benchmarking uses the dataset suggested by~\citet{RamirezGeffner2009}, or variations thereof~\cite{PereiraOrenMeneguzzi2020}. 
This dataset does not consider different sequences of observations for the same goal. 
This is not realistic since in the real world, agents under observation can generate different optimal plans for the same goal. 
A goal recogniser that does not consider this, risks jeopardising its accuracy. 
Therefore, in this paper we suggest a new measure for goal recognisers called \texttt{Resilience}, which measures the sensitivity of the goal recogniser to different observation sequences for the same goal hypothesis.
The contributions of this paper are: (1) a new dataset generation that eliminates the planner's bias and (2) a metric to measure the recogniser's resilience.

The paper first motivates the need to measure the resilience of a goal recogniser. Then presents background necessary for understanding the contribution for this paper, followed by the suggested dataset generation method. Finally, it evaluates the landmark-based recogniser's resilience, list conclusions and future work.

%% file: sections/02-motivating-example.tex
\begin{figure*}[!hpt]
    \centering
    \includegraphics[scale=0.65]{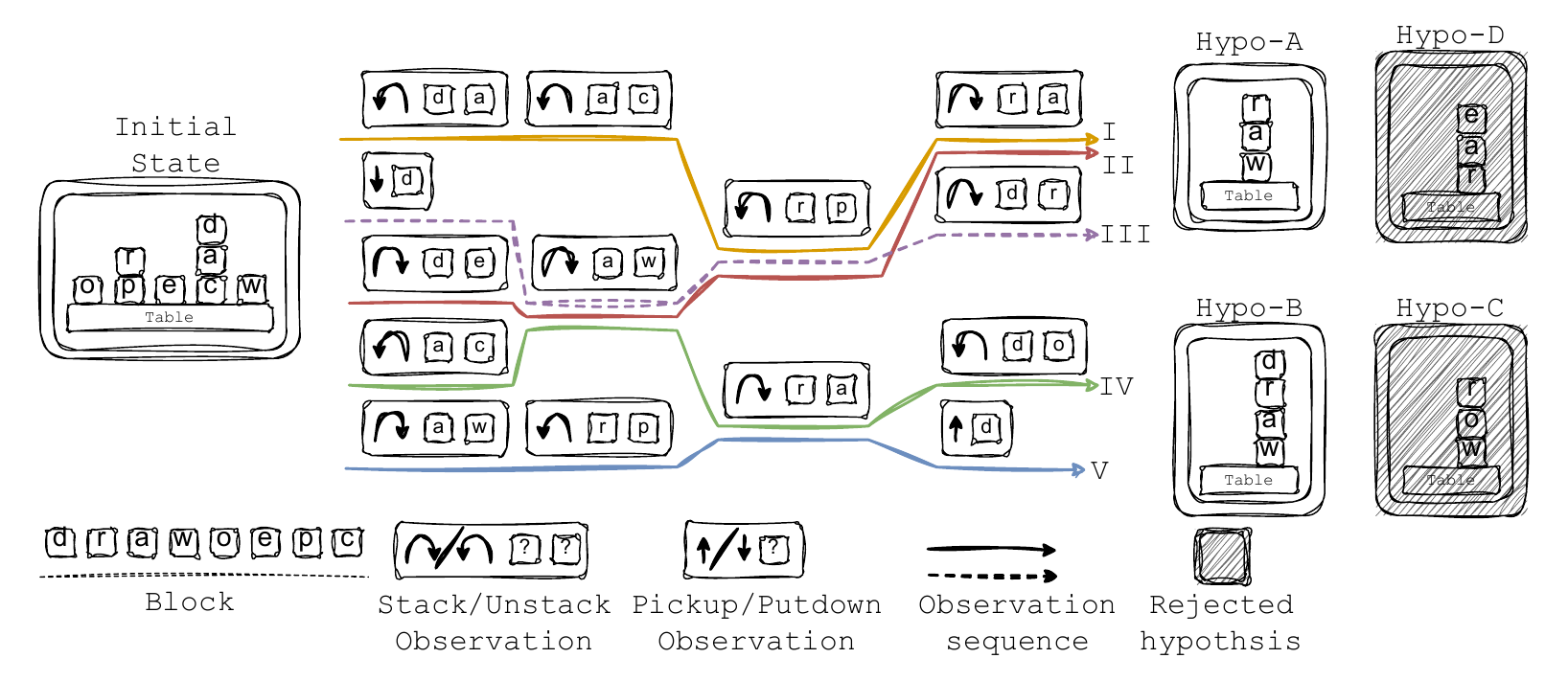}
    \caption{A complete goal recognition problem with a set of different observation sequences for the same blocks arrangement. }
    \label{fig:motivating-blocks-world-example}
\end{figure*}

\section{Motivating Example}\label{sec:motivating-example}

%

%
We illustrate the need for measuring a goal recogniser resilience through the well-known goal recognition task called \texttt{blocks-world}.
In this task, the recogniser attempts to infer the final placement of a set of lettered blocks from a partial sequence of observations. The proportion of the full sequence provided is called the \emph{observation level}.
\Cref{fig:motivating-blocks-world-example} shows a generated set of different observation sequences for the same arrangement. 
We refer to a specific target arrangement as a \emph{hypothesis}. 
This task has four hypotheses: (A) r on top a on top w, (B) d on top r on top a on top w, (C) r on top o on top w, and (D) e on top a on top r. Crucially, each hypothesis can typically be achieved through multiple distinct plans, each producing a different observation sequence. 
We generated 5 possible plans for hypothesis B, and then dropped 50\% of the actions randomly to create observation sequences with 50\% observability.
For this example, we used Symk~\cite{symk-with-expressive} to generate the observations and the Landmark-based recogniser~\cite{PereiraOrenMeneguzzi2020} to solve the tasks. 

Here, the landmark-based approach was only able to identify the correct hypothesis only one out of five times. 
Current benchmarking practice treats each observation sequence as an independent task, ignoring the fact that all five sequences represent different paths to the same goal.
We argue that a recogniser's performance on a hypothesis should instead be measured by aggregating results across all sequences for that hypothesis. For example, one could compute the proportion of correctly solved sequences and deem the recogniser resilient if this proportion exceeds a specified threshold.

%% file: sections/03-background.tex
\section{Background}
This section provides the necessary background for our suggested measurement. We first define the goal recognition task then cover top-k planners.



\subsection{Planning and Goal Recognition}



We formalise goal recognition in the context of automated planning~\cite{MeneguzziPereira2021}, starting with a high-level model and refining it to match the approaches discussed later. 
We then define goal recognition itself, grounding it in planning and progressively refining its components.

Following \citet{ghallab2016automated}, a planning task is a tuple of $\planningtask=\langle S, A, \gamma, \operatorname{cost}, I, G\rangle$, where the first part represents the domain with $S$ as a set of states, $A$ as a set of actions, and $\gamma$ as a transition function $\gamma: S \times A \rightarrow S$ that associates each state $s\in S$ and action $a\in A$ to the next state $\gamma(s,a)=s'$. The function $\operatorname{cost}:  A \rightarrow \mathbb{R}^+$ represents the cost of an action. Note that here we only consider action costs that are independent of the state unlike state-dependent action ones \cite{symk-with-expressive}.  $I\in S$ represents the initial state while $G$ is a formula that models all possible goal states.

A \textit{Goal Recognition} task is about identifying an agent's goal by observing its interactions with an environment~\cite[Chapter~1]{SukthankarGoldmanGeibEtAl2014}. 
These observations provide evidence for recognition and are either executed actions (\exemp, movement, cooking) or state changes (\exemp, location, activity). 
We adopt the problem formulation of~\citet{RamirezGeffner2009}, defining a goal recognition task (Definition~\ref{def:grtask}) analogously to a planning task but with two key differences. 
First, a set of \textit{goal hypotheses} $\goalconditions$ replaces the single goal. 
We assume there is a single \textit{correct} goal hypothesis $\goal^{*} \in \goalconditions$, which is not revealed to the recognizer.
Second, a sequence of observations $\observations$ reflects the agent's behaviour. 
Note that the goal state $G$ inside $\Xi$ is discarded. 
Thus, from this point we refer to $\Xi$ as the planning domain for simplicity. 

\begin{definition}[Goal Recognition Task]\label{def:grtask}  
	A goal recognition task $\Xi^\Omega_\mathcal{G} = \tuple{\planningdomain, \goalconditions, \observations}$ includes: a planning task $\planningdomain$; a set of possible goals $\goalconditions$; and an observation sequence $\observations$. 
\end{definition}




\subsection{Top-k planners}
\citet{riabov2014new} introduced the first top-k planner. It is called $\texttt{TK}^\ast$ and is based on the $\texttt{K}^\ast$ algorithm~\cite{aljazzar2011k}, which is a generalisation of $\texttt{A}^\ast$. This planner takes a solution to a planning task and creates a set of new planning tasks that preserve all solutions of the original task except for the specified one. It then searches for a reformulation tree, invoking an existing planner at each node. \citet{katz2018novel} proposed \texttt{FORBID-k}, an alternative iterative approach to top-k planning. Their method involves reformulating the planning task into a single task, allowing for the identification of additional solutions while preserving all previously found solutions except for the one originally given. They also suggested further methods to derive new solutions from those already discovered, which reduces the number of times the underlying cost-optimal planner needs to be invoked. Additionally, they extended their reformulation to prohibit multiple plans simultaneously, which helps mitigate the increase in the size of task formulations. \citet{speck2020symbolic} suggested using symbolic search for solving the top-k planning problem. They introduced \texttt{SymK}, a symbolic-based planner that performs better for small and large numbers of required plans compared to all other approaches. In this work, we used \texttt{SymK} because it is the state-of-the-art top-k planner.

%% file: sections/04-resilience-idea.tex
\section{Resilient Goal Recognition Dataset}
\label{sec:dataset}

In this section, we formulate a metric that measures the resilience of a goal recogniser, as discussed in motivating example. Furthermore, we show how to generate a resilient goal recognition dataset.

We define \emph{Version Coverage Score} (VCS) metric to measure the resilience of a goal recogniser for a given task. This metric is defined as follows:
\begin{definition}[Version Coverage Score (VCS)]\label{def:version_coverage_score}
Given a set of goal recognition tasks for the same hypothesis $\Gamma=\{\Xi^{\Omega_1}_\mathcal{G},\dots,\Xi^{\Omega_k}_\mathcal{G}\}$, a recogniser $\mathcal{R}$, Version Coverage Score is defined as $\operatorname{VCS}:\Gamma\times\{\mathcal{R}\}\rightarrow\mathbb{R}$ a function that receives a set of goal recognition tasks ($\Gamma$), a goal recogniser ($\mathcal{R}$) and computes the ratio between the correctly solved tasks to the number of tasks.
\end{definition}

After defining $\operatorname{VCS}$ metric, we say a goal recogniser $\mathcal{R}$ is resilient to a goal recognition task $\Xi_\mathcal{G}$ with $\Gamma$ tasks if its Version Coverage Score is greater than or equal to a resilience threshold $T \in [0,1]$ (i.e., $\operatorname{VCS}(\Gamma, \mathcal{R}) \geq T$).

In order to produce a resilient dataset we use a top-k planner~\cite{katz2018novel}. 
\Cref{alg:solver-behaviour} denotes a generalised procedure for generating such a dataset. 
Usually, a goal recognition dataset contains the same goal recognition task with different observation values $O\in\{10, 30, 50, 70, 100\}$ with noisy values $N\in\{0,10,20,30\}$~\cite{PereiraOrenMeneguzzi2020}. 
Furthermore, a goal recognition task can have at most $k$ distinct observations for a given hypothesis, resulting in a maximum of $k$ versions of the task. 
Therefore, to construct the whole dataset, we invoke \Cref{alg:solver-behaviour} for every observation, noise, $k$ and goal hypothesis.

\Cref{alg:solver-behaviour} receives a planning task ($\Xi$), planner ($\operatorname{planner}$), a goal hypothesis ($g$), the number of required variants of $\Xi$ ($k$), observability percentage ($O$) and noisy percentage ($N$). 
The first step of the planning task is updated to contain the goal hypothesis $g$ through an auxiliary function called $\operatorname{update}$, then generates $k$ observations using behaviour planning (lines \ref{line:task-update}-\ref{line:fbi-call}). 
In the second step, it generates a set of possible hypotheses $\mathcal{G}$ including the goal state in $\Xi$. 
This is done using the $\operatorname{HypothesisGenerator}$ function, which follows the same logic as the dataset from~\citet{RamirezG_AAAI2010}. 
Afterwards, for every ordered set of observations ($\Omega$), \Cref{alg:solver-behaviour} generates two tasks: a clean and a noisy version. 
This is achieved as follows, \Cref{alg:solver-behaviour} selects a subset of observations with size $O\cdot\vert\Omega\vert$ with no noise applied and then selects another subset of observations with size $O\cdot\vert\Omega\vert$ but with N\% noise applied (Lines~\ref{line:loop-start}-\ref{line:loop-end}). 

\begin{algorithm}
\caption{$\operatorname{TaskGenerator}$}\label{alg:solver-behaviour}
    \begin{algorithmic}[1]
        \REQUIRE $\Xi$: Planning task, $\operatorname{planner}$: Planner, $g$: goal hypothesis, $k$: planning task variants, $O$: observability percentage, $N$: noisy percentage.
        \ENSURE $k$ goal recognition tasks ($tasks_{clean}, tasks_{noise}$).
        \STATE $tasks_{clean}\gets\{\}, tasks_{noise}\gets\{\}$
        \STATE $\Xi^\prime\gets\operatorname{update}(\Xi,g)$\label{line:task-update}
        \STATE $\Psi\gets\operatorname{planner}(\Xi^\prime, k)$\label{line:fbi-call}
        \STATE $\mathcal{G}\gets\operatorname{HypothesisGenerator}(\Xi)\cup\{g\}$
        \FOR{$\Omega\in\Psi$}\label{line:loop-start}
            \STATE $\Omega^\prime_{clean}\gets\operatorname{Select}(\Omega, O, 0)$
            \STATE $\Omega^\prime_{noise}\gets\operatorname{Select}(\Omega, O, N)$
            \STATE $tasks_{clean}\gets tasks_{clean}\ \cup \{\langle\Xi,\mathcal{G},\Omega^\prime_{clean}\rangle\}$
            \STATE $tasks_{noise}\gets tasks_{noise}\ \cup \{\langle\Xi,\mathcal{G},\Omega^\prime_{noise}\rangle\}$
        \ENDFOR\label{line:loop-end}
        \RETURN $tasks_{clean}, tasks_{noise}$ \label{ret-stmt}
    \end{algorithmic}
\end{algorithm}

%% file: sections/05-exps.tex

\section{Experiments Setup, Results, \& Discussion}

\input{tables/grouped_symk_landmark}

In this section we generate a dataset using the method mentioned in the previous section, benchmarked the landmark-based goal recogniser on it and discussed the results of the experiment.

\paragraph{\textbf{Setup.}} To show resilience of a goal recogniser given different threshold levels, we generated a dataset with $k=5$ for every goal recognition task in \citet{RamirezGeffner2009}'s dataset using an Intel(R) Xeon(R) Gold 6138 CPU at 2.00GHz, with a 32GB RAM limit. 
Then we computed VCS for different resilience threshold values $T\in[0.1,1.0]$ the tasks' variants. 
As for the spread (i.e., number of selected hypotheses), we averaged them across the tasks' variants. 
Usually a goal recogniser is evaluated based on several parameters. 
The first parameter denotes the average number of goal hypotheses selected by the goal recogniser. 
Such parameter is called \textit{spread}. The second parameter is \textit{accuracy}, it is the number of correctly assigned truth-values to the goal hypotheses divided by
the total number of hypotheses. The last parameter is \textit{positive predictive value} (PPV),
it is the number of correctly selected true hypotheses divided by the total number of hypotheses selected as true.


\paragraph{\textbf{Results.}} To capture the relation between the observation level, resilience threshold and the evaluation metric (i.e., accuracy, PPV, spread) we tabled the statistical results showing the mean and standard deviation of each evaluation metric, see \Cref{tbl:results-overall} for the results.







\textbf{Discussion.}
In general, the values for accuracy and PPV decrease when the resilience threshold increases. 
This is expected because increasing the threshold means expecting that the recogniser can infer the correct hypothesis for more than one observation sequence.
Even though the landmark-based recogniser performs well for observation level above 50\%, its accuracy starts to deteriorate when the observation level is 50\% and below. 
This is not the under the old dataset. 
For instance on the old dataset, the recogniser had accuracy of 85\% at 10\% observation level and 95\% at 50\% observation level. 
This difference shows that even the state-of-the-art recogniser has a planner bias. 
We observe a similar pattern for the PPV. 
As for the spread, the same results matches the old dataset and the reason behind this is that the recogniser tend to select one hypothesis only.


%% file: tables/grouped_symk_landmark.tex
\begingroup
\setlength{\tabcolsep}{1.7pt}
\renewcommand{\arraystretch}{1.2}
\setlength{\intextsep}{2pt} 

\begin{table*}[t]
\centering

\begin{subtable}{0.95\textwidth}
\centering
\begin{tabular}{c|ccccccccccc}
\multirow{2}{*}{\begin{tabular}[c]{@{}c@{}}Obs. \\ Level\end{tabular}} & \multicolumn{11}{c}{Threshold} \\ 
 & 0.0 & 0.1 & 0.2 & 0.3 & 0.4 & 0.5 & 0.6 & 0.7 & 0.8 & 0.9 & 1.0 \\ \cline{1-12}
10 & 95.0/0.11 & 89.0/0.14 & 89.0/0.14 & 86.0/0.15 & 86.0/0.15 & 82.0/0.16 & 82.0/0.16 & 79.0/0.17 & 79.0/0.17 & 76.0/0.17 & 76.0/0.17 \\
30 & 97.0/0.05 & 96.0/0.07 & 96.0/0.07 & 94.0/0.09 & 94.0/0.09 & 92.0/0.11 & 92.0/0.11 & 89.0/0.12 & 89.0/0.12 & 85.0/0.14 & 85.0/0.14 \\
50 & 99.0/0.03 & 98.0/0.04 & 98.0/0.04 & 98.0/0.05 & 98.0/0.05 & 97.0/0.07 & 97.0/0.07 & 95.0/0.08 & 95.0/0.08 & 91.0/0.11 & 91.0/0.11 \\
70 & 99.0/0.02 & 99.0/0.03 & 99.0/0.03 & 99.0/0.03 & 99.0/0.03 & 98.0/0.05 & 98.0/0.05 & 98.0/0.07 & 98.0/0.07 & 96.0/0.09 & 96.0/0.09 \\
100 & 99.0/0.03 & 99.0/0.03 & 99.0/0.03 & 99.0/0.03 & 99.0/0.03 & 99.0/0.03 & 99.0/0.03 & 99.0/0.03 & 99.0/0.03 & 99.0/0.04 & 99.0/0.04 \\ \hline
\end{tabular}
\caption{Landmark accuracy statistical results (mean/std).}
\end{subtable}

\begin{subtable}{0.95\textwidth}
\centering
\begin{tabular}{c|ccccccccccc}
\multirow{2}{*}{\begin{tabular}[c]{@{}c@{}}Obs. \\ Level\end{tabular}} & \multicolumn{11}{c}{Threshold} \\ 
 & 0.0 & 0.1 & 0.2 & 0.3 & 0.4 & 0.5 & 0.6 & 0.7 & 0.8 & 0.9 & 1.0 \\ \cline{1-12}
10 & 81.0/0.37 & 57.0/0.43 & 57.0/0.43 & 45.0/0.45 & 45.0/0.45 & 31.0/0.4 & 31.0/0.4 & 22.0/0.35 & 22.0/0.35 & 13.0/0.29 & 13.0/0.29 \\
30 & 85.0/0.18 & 78.0/0.29 & 78.0/0.29 & 70.0/0.36 & 70.0/0.36 & 61.0/0.42 & 61.0/0.42 & 49.0/0.45 & 49.0/0.45 & 35.0/0.44 & 35.0/0.44 \\
50 & 91.0/0.14 & 90.0/0.19 & 90.0/0.19 & 86.0/0.26 & 86.0/0.26 & 80.0/0.34 & 80.0/0.34 & 70.0/0.41 & 70.0/0.41 & 54.0/0.47 & 54.0/0.47 \\
70 & 95.0/0.12 & 95.0/0.13 & 95.0/0.13 & 93.0/0.17 & 93.0/0.17 & 90.0/0.24 & 90.0/0.24 & 87.0/0.31 & 87.0/0.31 & 77.0/0.4 & 77.0/0.4 \\
100 & 95.0/0.16 & 95.0/0.17 & 95.0/0.17 & 95.0/0.17 & 95.0/0.17 & 95.0/0.17 & 95.0/0.17 & 95.0/0.18 & 95.0/0.18 & 93.0/0.21 & 93.0/0.21 \\ \hline
\end{tabular}
\caption{Landmark ppv statistical results (mean/std).}
\end{subtable}

\begin{subtable}{0.95\textwidth}
\centering
\begin{tabular}{c|ccccccccccc}
\multirow{2}{*}{\begin{tabular}[c]{@{}c@{}}Obs. \\ Level\end{tabular}} & \multicolumn{11}{c}{Threshold} \\ 
 & 0.0 & 0.1 & 0.2 & 0.3 & 0.4 & 0.5 & 0.6 & 0.7 & 0.8 & 0.9 & 1.0 \\ \cline{1-12}
10 & 1.47/0.73 & 1.47/0.73 & 1.47/0.73 & 1.47/0.73 & 1.47/0.73 & 1.47/0.73 & 1.47/0.73 & 1.47/0.73 & 1.47/0.73 & 1.47/0.73 & 1.47/0.73 \\
30 & 1.26/0.41 & 1.26/0.41 & 1.26/0.41 & 1.26/0.41 & 1.26/0.41 & 1.26/0.41 & 1.26/0.41 & 1.26/0.41 & 1.26/0.41 & 1.26/0.41 & 1.26/0.41 \\
50 & 1.13/0.27 & 1.13/0.27 & 1.13/0.27 & 1.13/0.27 & 1.13/0.27 & 1.13/0.27 & 1.13/0.27 & 1.13/0.27 & 1.13/0.27 & 1.13/0.27 & 1.13/0.27 \\
70 & 1.08/0.2 & 1.08/0.2 & 1.08/0.2 & 1.08/0.2 & 1.08/0.2 & 1.08/0.2 & 1.08/0.2 & 1.08/0.2 & 1.08/0.2 & 1.08/0.2 & 1.08/0.2 \\
100 & 1.11/0.39 & 1.11/0.39 & 1.11/0.39 & 1.11/0.39 & 1.11/0.39 & 1.11/0.39 & 1.11/0.39 & 1.11/0.39 & 1.11/0.39 & 1.11/0.39 & 1.11/0.39 \\ \hline
\end{tabular}
\caption{Landmark spread statistical results (mean/std).}
\end{subtable}

\caption{Statistical results for accuracy, ppv and spread. Results format is mean/std.}\label{tbl:results-overall}
\end{table*}

\endgroup

%% file: sections/06-conclusion-future-work.tex
\section{Conclusions \& Future work}
\label{sec:conclusions}

This paper addresses a critical limitation in existing goal recognition datasets which is related to the systematic planner bias. 
Current datasets do not consider that optimal planners can generate different plans for the same planning task and this what introduces the bias. 
To overcome this limitation, we developed a new method to generate datasets that produces multiple plans for each goal hypothesis. 
This can be done by leveraging top-k planners such as SymK. 
We proposed a metric to quantify a goal recogniser's resilience and called it Version Coverage Score (VCS). 
Our experimental evaluation of the state-of-the-art landmark-based goal recogniser on this new dataset revealed significant insights. 
While the recogniser performs well under high observation levels, its accuracy and positive predictive value deteriorate as the observation level decreases and the resilience threshold increases. 
This decline, not observed when testing on the traditional biased dataset and indicates that even state-of-the-art recogniser is susceptible to planner bias. 
5
There are several promising directions emerge from this research. 
One of these directions is to explore using diverse planners to generate more challenging tasks for goal recognisers. 
Another possible direction is exploring other methods to pick which actions of the observation sequence to drop or modify. 
Currently, it is based on a random sampling approach.